\documentclass[journal,10pt]{IEEEtran}
\usepackage{cite}
\usepackage{amsmath,amssymb,amsfonts}
\usepackage{algorithm,algorithmic}
\usepackage{graphicx}
\usepackage{textcomp}
\usepackage{dblfloatfix}
\usepackage{pdfpages} 
\usepackage[hidelinks]{hyperref}
\usepackage{etoolbox}

\def\BibTeX{{\rm B\kern-.05em{\sc i\kern-.025em b}\kern-.08em
    T\kern-.1667em\lower.7ex\hbox{E}\kern-.125emX}}

\begin{document}
\title{Context-Continuous Preference Learning for Exoskeleton Personalization}
\author{Sunin Baek, Sungwoo Park, and Daekyum Kim
\thanks{This work was supported by the Korean ARPA-H Project through the Korea Health Industry Development Institute (KHIDI), funded by the Ministry of Health and Welfare, Republic of Korea (RS-2025-25455839), and by the National Research Foundation of Korea (NRF), funded by the Ministry of Science and ICT, Republic of Korea (RS-2026-25499546). }
\thanks{Sunin Baek is with the School of Mechanical Engineering, Korea University, Seoul 02841, South Korea (e-mail: suninbaek@korea.ac.kr).}
\thanks{Sungwoo Park is with the Innovation Lab, WIRobotics, Yongin 16942, South Korea (e-mail: sungwoop@wirobotics.com).}
\thanks{Daekyum Kim is with the School of Mechanical Engineering and the School of Smart Mobility, Korea University, Seoul 02841, South Korea (e-mail: daekyum@korea.ac.kr).}
\thanks{Corresponding authors: Sungwoo Park and Daekyum Kim.}
\thanks{This work has been submitted to the IEEE for possible publication. Copyright may be transferred without notice, after which this version may no longer be accessible.}
}

\maketitle

\begin{abstract}
Personalizing exoskeleton assistance across operating conditions is constrained by the time and physical effort required to collect user feedback. We examined whether a user's preference landscape varies smoothly across operating conditions and when this continuity supports learning from limited feedback. We propose Context-Continuous Preference Learning (CCPL), a Gaussian-process preference model that shares observations across nearby contexts while retaining context-specific utility estimates. We evaluated CCPL through simulations and retrospective analyses of ankle and elbow exoskeleton preference data from nine healthy adults. In simulations, CCPL improved reconstruction and preference-based Bayesian optimization relative to independent learning when preferences varied smoothly, but showed negative transfer when continuity was weak. In both human studies, full-data reference landscapes estimated separately for each participant and context tended to be more similar between nearby operating conditions. With five exposures per context, CCPL increased mean reconstruction correlation with these references from 0.644 to 0.720 for ankle assistance and from 0.476 to 0.526 for elbow assistance relative to independent learning. The five-exposure budget was approximately 37\% lower for ankle and 17\% lower for elbow than the estimated independent-learning budgets needed to match these correlations. CCPL also improved held-out response prediction relative to independent learning, while benefits over pooled learning varied. These findings support context continuity as a basis for sharing preference observations under limited feedback, although benefits for online personalization in humans remain to be established.
\end{abstract}


\section{Introduction}
\IEEEPARstart{E}{xoskeletons} can support or augment human movement across lower- and upper-limb tasks \cite{ref1}. Users differ in how they respond and adapt to assistance, so realizing these benefits often requires tailoring assistance to the individual \cite{ref2,ref3}. Human-in-the-loop methods personalize assistance by updating controller parameters using physiological, biomechanical, task-performance, or user-reported measures \cite{ref1,ref4}. Such methods have used metabolic cost for lower-limb assistance \cite{ref2,ref5} and electromyographic activity for upper-limb assistance \cite{ref6,ref7,ref8}. However, obtaining user-specific measurements or feedback requires the user to experience the assistance settings being evaluated and may also require additional equipment, such as metabolic measurement systems or electromyography sensors along with stable observation periods or repeated trials \cite{ref9,ref10}. Session duration, fatigue, adaptation, and safety can therefore limit the amount of informative user-specific data available for personalization \cite{ref3,ref11,ref12}.

This data constraint becomes more consequential when assistance must be personalized across operating contexts, such as walking at different speeds and lifting different loads. Lower-limb studies have addressed such context dependence through condition-specific optimization \cite{ref13}, measurements of individual muscle dynamics \cite{ref14}, interpolation of optimized parameters across walking speeds \cite{ref15}, and learned relationships between optimized assistance and gait conditions \cite{ref16}. Similarly, studies of upper-limb and back-support assistance have learned assistive policies that generalize across tasks \cite{ref17} and adapted assistance to changes in movement or loading demands \cite{ref6,ref7,ref18,ref19,ref20}. These approaches highlight the importance of accounting for changes in assistance requirements across operating conditions, including user-specific differences.
For example, individualized ankle assistance profiles have been derived from measurements of each user’s soleus muscle dynamics at different walking speeds and inclines \cite{ref14}.
When these measurements are collected separately for each walking condition, adding conditions requires further observations from the same user.
This scaling challenge motivates reusing observations collected from the same user in other contexts, while preserving context-specific differences in personalized assistance.

The need for information reuse across contexts also extends to preference-based personalization, in which users’ assessments of assistance guide the selection of controller parameters. These assessments can reflect comfort, perceived support, and other aspects of the user's experience \cite{ref21,ref22}. Users can express their preferences through direct self-tuning \cite{ref23,ref24}, ordinal ratings \cite{ref25}, or pairwise comparisons \cite{ref21,ref22}. Ordinal and pairwise responses can be used to estimate latent utility functions from noisy feedback and guide preference-based Bayesian optimization (PBO) \cite{ref22,ref26,ref27,ref28}. We refer to a utility function representing a user's relative preferences over assistance parameters as the user's preference landscape. Preference-based approaches have been applied to ankle and hip assistance, exoskeleton gait design, lifting, and upper-limb assistance \cite{ref21,ref22,ref25,ref29,ref30,ref31,ref32}. Although preference feedback can reduce reliance on physiological measurements, evaluating each setting still requires physical exposure and a user judgment \cite{ref22}.

Sharing preference observations across operating contexts may reduce the need to collect new feedback in each condition. In the broader machine learning literature, contextual bandit methods incorporate context when learning from scalar rewards or pairwise comparisons \cite{ref33,ref34}, while multitask models share information across related optimization or preference-learning tasks \cite{ref35,ref36}. Such sharing has also been implemented in context-aware Gaussian-process preference models for image color grading \cite{ref37}. For exoskeleton personalization, the value of sharing depends on whether observations from one condition remain informative about preferences in another. Exoskeleton studies have optimized hip assistance separately at different walking speeds using preference feedback \cite{ref38} and used population-derived Gaussian-process models incorporating user preference to select back-support assistance across lifting loads \cite{ref39}. However, it remains unclear how an individual's preference landscapes are related across conditions and when sharing observations between those conditions improves or impairs learning.

To examine these relationships, we focus on continuous operating variables such as walking speed or lifting load. Muscle-based assistance profiles vary with walking speed \cite{ref14}, and muscle activity around the elbow joint varies with external torque \cite{ref40}. Changes in these conditions may alter how comfortable and supportive a given assistance profile feels to the user. We hypothesize that an individual's preference landscape varies smoothly with operating context, such that nearby conditions tend to have more similar preference landscapes than more distant conditions. We use \emph{context continuity} to describe the extent of this smooth variation at the scale of the sampled conditions. A Gaussian-process preference model can encode this smoothness assumption through its covariance function, enabling information sharing between nearby conditions \cite{ref33,ref37}. When this assumption holds, sharing may improve preference estimates with limited feedback in the target condition. If preferences differ more across conditions than the model assumes, sharing may instead lead to negative transfer relative to learning each context separately  \cite{ref41}.

This study examines whether a user’s preference landscape varies smoothly across operating conditions and when this continuity supports learning from limited feedback. 
We introduce Context-Continuous Preference Learning (CCPL), which represents an individual's preference landscape as a latent utility function defined jointly over assistance parameters and operating context. 
Its Gaussian-process preference model shares information between nearby conditions while retaining context-specific preference estimates.  
We compare CCPL with independent preference learning, which models each context separately, and pooled preference learning, which combines observations without retaining context.
In simulations, we vary context continuity, action and context dimensionality, and feedback budgets to examine when sharing improves latent utility reconstruction and PBO performance and when it leads to negative transfer. 
We also collect preference data in human exoskeleton experiments that vary walking speed for ankle assistance and external load and fatigue-induction dose for elbow assistance. 
Retrospective analyses assess evidence for context continuity in these data and evaluate reconstruction from limited feedback, prediction of held-out preference responses, and the value of reusing observations from other operating contexts.

\section{Methods}

\subsection{Context-Continuous Preference Learning Framework}

\subsubsection{Problem Formulation and Preference Observations}

We consider preference-based personalization for an individual user across measured operating contexts. For user $s$, let $\mathbf{x}\in\mathcal{X}\subset\mathbb{R}^{d_x}$ denote an admissible vector of assistance parameters, hereafter called an action, and let $\mathbf{c}\in\mathcal{C}\subset\mathbb{R}^{d_c}$ denote the measured operating context. The user's preference landscape is represented by a latent utility function $f_s:\mathcal{X}\times\mathcal{C}\rightarrow\mathbb{R}$, where larger values indicate greater preference among actions evaluated in the same context. Fixing $\mathbf{c}$ gives a context-specific slice $f_s(\cdot,\mathbf{c})$ of this joint landscape. The learning objectives are to reconstruct relative utility over the action space and identify a preferred action at each context,
\begin{equation}
\mathbf{x}_s^*(\mathbf{c})=\underset{\mathbf{x}\in\mathcal{X}}{\arg\max}\,f_s(\mathbf{x},\mathbf{c}).\label{eq:optimal_action}
\end{equation}
Models are fitted separately for each user, and the user subscript is omitted hereafter for notational simplicity.

Preference observations comprise ordinal assessments of individual actions and direct pairwise responses comparing two actions experienced in the same context. Ordinal assessments classify individual actions as unacceptable, ambiguous, or acceptable and yield comparisons between differently ranked actions. Direct responses indicate preference for either action or a tie. All comparisons remain within their observed contexts and constrain utility differences without directly identifying absolute utility offsets between contexts.

Feedback budgets count assistance exposures, each evaluating one action in one context.
Ordinal-derived comparisons are constructed from existing assessments and require no additional user exposure or feedback.

\subsubsection{Gaussian-Process Preference Model}
\label{sec:gp_model}
We model latent utility using a Gaussian-process (GP) preference formulation \cite{ref27}. Let $\mathbf{z}\in\mathbb{R}^{d_z}$ denote the model input, whose components depend on the learning strategy described in Section~\ref{sec:learning_strategies}. All input dimensions are normalized to [0,1] using the bounds of the evaluated domain. For $n$ distinct model inputs, let $\mathbf{f}=[f(\mathbf{z}_1),\ldots,f(\mathbf{z}_n)]^\top$ denote the corresponding latent utilities. Repeated observations at the same model input refer to the same latent variable. We use a zero-mean GP prior, $\mathbf{f}\sim\mathcal{N}(\mathbf{0},\mathbf{K})$, with covariance
\begin{equation}
K_{jk}=\sigma_f^2\exp\left[-\frac{1}{2}\sum_{h=1}^{d_z}\left(\frac{z_{jh}-z_{kh}}{\ell_h}\right)^2\right]+\eta\delta_{jk},\label{eq:gp_covariance}
\end{equation}
where $z_{jh}$ is the $h$th coordinate of input $\mathbf{z}_j$, $\ell_h$ is the corresponding length scale, $\sigma_f^2$ is the signal variance, and $\eta$ is the diagonal regularization parameter. Here, $\delta_{jk}$ equals one when $j=k$ and zero otherwise.

For comparison $i$, define the utility difference as $\Delta_i=f(\mathbf{z}_i^A)-f(\mathbf{z}_i^B)$. The response $y_i\in\{1,0,-1\}$ indicates preference for action A, a tie, or preference for action B, respectively. Let $\sigma_p$ denote the standard deviation of noise on the utility difference and $\epsilon$ the indifference threshold. We use a probit preference likelihood \cite{ref27} with an indifference interval for ties \cite{ref42}. The resulting three-category response probabilities follow an ordinal-threshold formulation \cite{ref26}:
\begin{subequations}
\label{eq:response_probabilities}

\begin{equation}
p(y_i=1\mid\Delta_i)=
\Phi\!\left(\frac{\Delta_i-\epsilon}{\sigma_p}\right),
\label{eq:response_positive}
\end{equation}

\begin{equation}
p(y_i=0\mid\Delta_i)=
\Phi\!\left(\frac{\epsilon-\Delta_i}{\sigma_p}\right)
-\Phi\!\left(\frac{-\epsilon-\Delta_i}{\sigma_p}\right),
\label{eq:response_tie}
\end{equation}

\begin{equation}
p(y_i=-1\mid\Delta_i)=
\Phi\!\left(\frac{-\Delta_i-\epsilon}{\sigma_p}\right).
\label{eq:response_negative}
\end{equation}

\end{subequations}
where $\Phi$ is the standard normal cumulative distribution function. Directly reported ties are retained during fitting and favor utility differences within the indifference interval $[-\epsilon,\epsilon]$.

Direct pairwise responses receive unit weight. To construct ordinal-derived comparisons, we use each ordinal assessment as an anchor $r$. Eligible partners are observations of different actions in the same context with unequal ordinal ranks. Action pairs already represented by a direct response in that context are excluded. Let $\mathcal{P}_r$ denote the selected eligible partners and $m_r=|\mathcal{P}_r|$. When $m_r>0$, each derived comparison favors the action with the higher ordinal rank and receives weight
\begin{equation*}
w_{rj}^{\mathrm{ord}}=\frac{0.5}{m_r},\qquad j\in\mathcal{P}_r.
\end{equation*}
Each contributing anchor generates a total comparison weight of 0.5, half that of a direct pairwise response, regardless of the number of eligible partners. Ordinal-derived comparisons and their weights are rebuilt from the observations available at each model fit.

Let $\mathcal{D}=\{(\mathbf{z}_i^A,\mathbf{z}_i^B,y_i,w_i)\}_{i=1}^{M}$ denote the weighted comparison records. We use a weighted-likelihood approximation to combine direct responses with ordinal-derived comparisons. Both comparison types use the response probabilities in \eqref{eq:response_probabilities} with the same likelihood parameters. The corresponding negative log posterior, up to an additive constant, is
\begin{equation}
g(\mathbf{f})=\frac{1}{2}\mathbf{f}^{\top}\mathbf{K}^{-1}\mathbf{f}-\sum_{i=1}^{M}w_i\log p(y_i\mid\Delta_i).\label{eq:posterior}
\end{equation}
We obtain the posterior mode by minimizing \eqref{eq:posterior} and approximate the posterior as Gaussian using the Laplace method \cite{ref27}. Predictive means and covariances at query inputs are computed by integrating the GP conditional distribution over this approximation. The regularization term $\eta$ is included in the training covariance matrix. Cross-covariances and covariances at query inputs use the signal kernel.

Kernel and likelihood hyperparameters are fixed across users and limited-feedback subsets in the primary analyses. All primary models use $\sigma_f^2=1$, $\sigma_p=0.20$, $\epsilon=0.15$, and $\eta=10^{-4}$. Study-specific length scales and ordinal-partner selection settings are specified in Sections~\ref{sec:simulation} and~\ref{sec:human_studies}.

\subsubsection{Learning Strategies and Comparators}
\label{sec:learning_strategies}
\begin{figure*}[!t]
\centerline{\includegraphics[width=\textwidth]{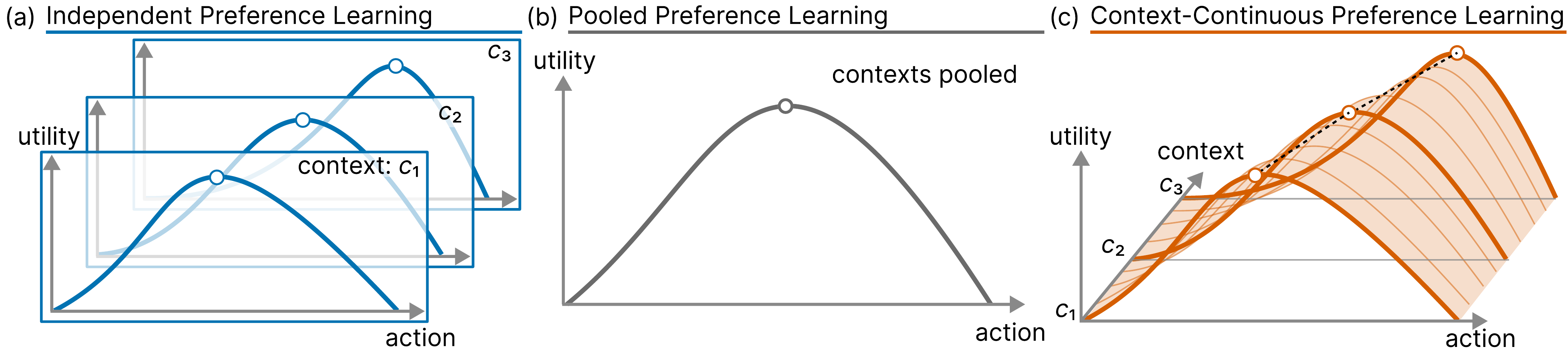}}
\caption{Preference-learning strategies across contexts. Comparisons remain within their observed contexts. (a) Independent learning fits a separate utility function over actions for each context. (b) Pooled learning combines observations into one utility function without context as an input. (c) CCPL models utility over action and context, sharing information through covariance while retaining context-specific predictions. White circles mark utility maxima. Thin orange curves show utility slices at intermediate contexts, and the dashed curve traces the maximizing action across contexts.}
\label{fig:1}
\end{figure*}

Let $\mathcal{D}_m$ denote the comparison set for context $\mathbf{c}_m$.
The three strategies share the observation representation, likelihood, posterior approximation, action normalization, and action covariance settings within each study.
They differ in their use of context (Fig.~\ref{fig:1}).

Independent preference learning fits a separate GP $f_m(\mathbf{x})$ over actions for each context, using input $\mathbf{z}=\mathbf{x}$ and conditioning only on $\mathcal{D}_m$. Identical actions across contexts have separate latent utilities, allowing different predictions.

Pooled preference learning fits a single GP $f_{\mathrm{pool}}(\mathbf{x})$ over actions across contexts, using input $\mathbf{z}=\mathbf{x}$ and combining within-context comparison sets without introducing cross-context comparisons. Identical actions share a latent utility, so predictions cannot vary across contexts.

Context-Continuous Preference Learning (CCPL) fits a single GP $f(\mathbf{x},\mathbf{c})$ over actions and contexts, using input $\mathbf{z}=[\mathbf{x}^{\top},\mathbf{c}^{\top}]^{\top}$ and conditioning on the combined within-context comparison sets. Identical actions across contexts have distinct latent utilities linked by context covariance, allowing context-specific predictions. With separate length scales for action and context dimensions, the signal covariance in \eqref{eq:gp_covariance} factorizes into action and context components \cite{ref33,ref37}:
\begin{equation}
k_{\mathrm{CCPL}}\big((\mathbf{x},\mathbf{c}),(\mathbf{x}',\mathbf{c}')\big)=\sigma_f^2k_x(\mathbf{x},\mathbf{x}')k_c(\mathbf{c},\mathbf{c}'),\label{eq:ccpl_kernel}
\end{equation}
where $k_x$ and $k_c$ are unit-variance squared-exponential covariance functions over normalized action and context coordinates, respectively. The diagonal regularization in \eqref{eq:gp_covariance} is retained in the training covariance matrix.

Context length scales control how quickly correlation decreases with differences in context variables. Short scales weakly couple distinct contexts, whereas long scales encourage similar preferences. Comparisons with independent learning assess the value of cross-context observations, while comparisons with pooled learning assess the value of retaining their context.

\subsection{Simulation Study}
\label{sec:simulation}
\subsubsection{Simulation Setup}

We constructed synthetic preference-learning problems over the normalized action domain $\mathcal{X}=[0,1]^{d_x}$ and context domain $\mathcal{C}=[0,1]^{d_c}$. Action dimensionality was varied over $d_x\in\{2,3,4\}$, with 10 equally spaced grid values per dimension, yielding 100, 1,000, or 10,000 admissible actions. Context dimensionality was varied over $d_c\in\{1,2\}$, with five equally spaced values per dimension, yielding 5 or 25 sampled contexts. We generated 30 independent instances of the synthetic preference-learning problem for each of the six dimensionality configurations, giving 180 instances in total. For each instance, all three strategies were evaluated on the same preference landscapes. Random seeds were assigned according to a fixed rule for reproducibility.

For each instance, we generated a smoothly varying component using a randomly rotated, anisotropic plateau-shaped utility function with a small smooth sinusoidal residual. Its center, axis-specific widths, decay rate, plateau extent, and shape exponent varied with context through polynomial functions within specified bounds. Independent context-specific components used the same functional form with parameters sampled separately for each context. Supplementary Section B1 provides the complete generator, parameter distributions, bounds, and context-continuity check.

Context continuity was controlled by mixing the smoothly varying and independently generated utility components. At context $\mathbf{c}_m$, the ground-truth utility was
\begin{equation}
u_{\gamma}(\mathbf{x},\mathbf{c}_m)=\mathcal{N}_m\left[\gamma u_{\mathrm{sm}}(\mathbf{x},\mathbf{c}_m)+(1-\gamma)u_{\mathrm{ind}}(\mathbf{x},\mathbf{c}_m)\right].\label{eq:synthetic_utility}
\end{equation}
Here, $u_{\mathrm{sm}}$ and $u_{\mathrm{ind}}$ denote the smoothly varying and independently generated components, respectively. Each component was min-max normalized over the action grid within each context, and $\mathcal{N}_m$ rescales their mixture to [0,1] within context $m$. We evaluated $\gamma\in\{0,0.25,0.50,0.75,1\}$. The same component landscapes were used across all five $\gamma$ settings within each instance. At $\gamma=0$, utilities were generated independently across contexts, whereas $\gamma=1$ retained only the smoothly varying component. This manipulation changed the underlying preference structure while the fitted model settings remained fixed.

Ordinal assessments and direct pairwise responses were generated from normalized ground-truth utility. For ordinal assessments, two utility thresholds separated the unacceptable, ambiguous, and acceptable categories. The lower and upper thresholds were sampled uniformly from [0.25,0.40] and [0.60,0.75], respectively, for each instance and retained across contexts. To introduce ordinal response noise, each ordinal label was independently changed to a uniformly selected adjacent category with probability 0.02. For direct pairwise responses, the within-context utility difference was defined as $\Delta=u_{\gamma}(\mathbf{x}^{A},\mathbf{c})-u_{\gamma}(\mathbf{x}^{B},\mathbf{c})$. A tie was generated when $|\Delta|\leq0.15$. Outside this interval, action A was preferred with probability $\Phi(\Delta/0.20)$, and action B otherwise. Thus, ties were deterministic within this interval, whereas the fitted likelihood in Section~\ref{sec:gp_model} allowed probabilistic ties.

Reconstruction and PBO used seven feedback budgets per context,
$B\in\{5,10,15,20,30,40,50\}$, yielding $B$ ordinal assessments
and $B-1$ direct comparisons between consecutive actions. Normalized action length scales were fixed at 0.30 for all models, and normalized context length scales were fixed at 0.50 for CCPL. To limit the number of ordinal-derived comparisons, each anchor retained at most 32 eligible partners, sampled uniformly without replacement when more were available. Other model settings followed Section~\ref{sec:gp_model}.

\begin{figure*}[!b]
\centerline{\includegraphics[width=\textwidth]{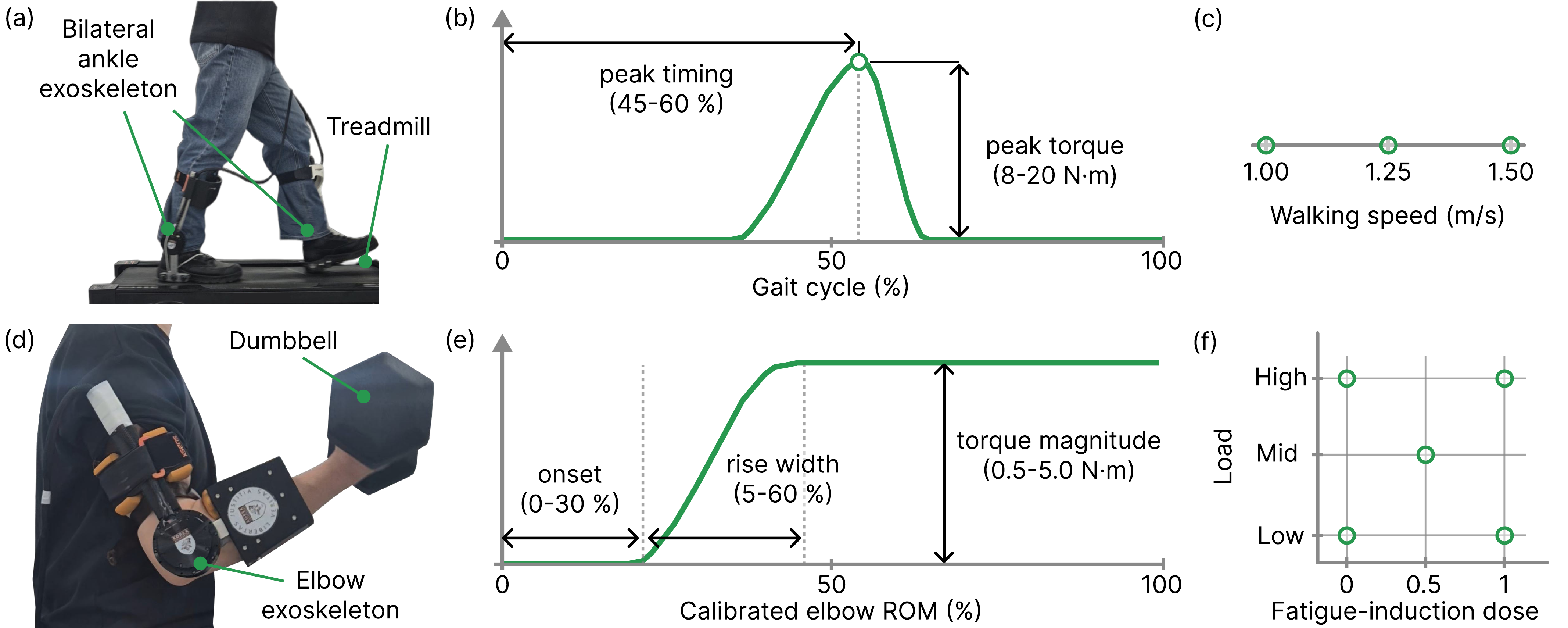}}
\caption{Human experimental systems, assistance profiles, and contexts. (a) Bilateral ankle exoskeleton for treadmill walking. (b) Ankle peak plantarflexion torque and peak timing. (c) Walking speeds of 1.00, 1.25, and 1.50 m/s. (d) Elbow exoskeleton for loaded flexion. (e) Elbow torque magnitude, onset position, and rise width over the participant-specific calibrated range of motion (ROM). (f) Five contexts combining participant-specific low and high loads with fatigue-induction doses of 0 and 1, plus intermediate load at dose 0.5. Dose is prescribed unassisted curls divided by 30.}
\label{fig:2}
\end{figure*}

\subsubsection{Evaluation Protocols}

For limited-feedback reconstruction, we generated a new action sequence for each instance, context-continuity setting, context, and feedback budget. Actions were selected without using preference responses, and all strategies received the same action sequence, ordinal assessments, and direct pairwise responses. Each sequence began at the grid point nearest the center of the normalized action domain, followed by uniformly sampled actions subject to distance and recent-action constraints. Reconstruction was evaluated by Pearson correlation between the posterior mean and ground-truth utility over the complete action grid within each context, with constant predictions assigned a correlation of zero.

For PBO, each strategy followed its own adaptive trajectory on the same ground-truth landscapes.
The first exposure in each context used the grid point nearest the center of the normalized action domain and yielded only an ordinal assessment.
Subsequent actions were sampled randomly until the model had comparison data, then selected by line-restricted posterior sampling \cite{ref30}.

Each subsequent exposure yielded an ordinal assessment and a direct comparison with the preceding action in that context. After every exposure, ordinal-derived comparisons were rebuilt, and the model was updated using the rebuilt comparisons and any available direct responses, including ties. Runs continued to 50 exposures per context. At each specified budget, the action maximizing the posterior mean over the complete grid was recommended and evaluated by its normalized ground-truth utility. Action-selection details for both reconstruction and PBO are provided in Supplementary Section B1.

The dimensionality analysis compared reconstruction and optimization across the six configurations at $\gamma=1$ and 10 exposures per context. The context-continuity analysis compared budget-averaged performance across the five $\gamma$ settings. For each instance and method, budget-averaged performance was the trapezoidal area under the performance curve from 5 to 50 exposures, divided by the budget span of 45 exposures. Negative transfer was defined as lower CCPL performance than independent learning under the same evaluation condition.

\subsection{Human Studies}
\label{sec:human_studies}
\subsubsection{Participants, Ethics, and General Procedures}

Nine healthy adults participated in two studies of ankle assistance during treadmill walking and elbow assistance during loaded flexion. Six completed the ankle study, six completed the elbow study, and three participated in both. The study protocol was approved by the Korea University Institutional Review Board (KUIRB-2026-0475-01), and all participants provided written informed consent before participation. Participant characteristics and participation are in Supplementary Table S1.

After each exposure, participants reported an ordinal assessment of unacceptable, ambiguous, or acceptable to the experimenter, using ambiguous when neither judgment was clear. From the second exposure in each context, they compared the current and preceding settings, with ties permitted. If a direct response favored the lower-rated setting, participants repeated the current ordinal assessment and pairwise response until the conflict was resolved. Only the final responses after this consistency check were modeled. Assistance sequences were independent of responses, and all strategies were evaluated retrospectively on the same recorded histories.

\subsubsection{Ankle Exoskeleton Study}

The ankle study used a custom-built bilateral exoskeleton (Fig.~\ref{fig:2}(a)) with two AK70-10 KV100 actuators (CubeMars, China). Inertial sensors on each foot's dorsum detected heel strikes, and gait phase was estimated from the time since the latest strike and recent stride durations. Desired torques were converted to motor-current commands updated at 200 Hz. Peak plantarflexion torque ranged from 8 to 20 N\textperiodcentered{}m in 0.5 N\textperiodcentered{}m increments, and peak timing ranged from 45\% to 60\% of the gait cycle in increments of 0.5 percentage points (Fig.~\ref{fig:2}(b)). Rise and fall intervals were fixed at 20\% and 8\% of the gait cycle, respectively, with cubic interpolation and zero endpoint slopes. These ranges and intervals were selected with reference to previous ankle exoskeleton studies to explore preferences across assistance magnitudes and timings \cite{ref2,ref15,ref23}. The action grid contained 775 admissible profiles.

Participants evaluated 101 distinct assistance profiles at each of three walking speeds (1.00, 1.25, and 1.50 m/s) on a motorized treadmill (Fig.~\ref{fig:2}(c)). The speed order was randomized for each participant, with at least 5 minutes of rest between speed conditions to limit carryover \cite{ref2}. Participants were allowed to extend the rest period as needed before continuing \cite{ref24}. For ordinal assessments, participants judged the acceptability of the assistance for continued walking \cite{ref25}. Exposure duration was not fixed, and participants provided feedback when they felt they had sufficiently experienced each setting. The initial nine profiles at each speed formed a 3 $\times$ 3 grid using each action variable's minimum, midpoint, and maximum. The center profile was presented first, followed by the remaining eight in random order. Subsequent profiles were sampled uniformly from settings not yet evaluated at that speed and satisfying minimum-distance constraints relative to the two preceding settings. Each speed context yielded 101 ordinal assessments and 100 direct pairwise responses.

\subsubsection{Elbow Exoskeleton Study}

The elbow study used a custom-built single-joint exoskeleton (Fig.~\ref{fig:2}(d)) with one AK70-10 KV100 actuator (CubeMars, China). The actuator encoder tracked elbow flexion, and desired torques were converted to motor-current commands updated at 200 Hz. Before testing in each context, participants performed five unloaded curls to calibrate the range of motion (ROM) for assistance control. Torque magnitude ranged from 0.5 to 5.0 N\textperiodcentered{}m in 0.5 N\textperiodcentered{}m increments. Onset position ranged from 0\% to 30\% ROM, and rise width from 5\% to 60\% ROM, both in increments of five percentage points (Fig.~\ref{fig:2}(e)). These ranges were selected to explore preferences across assistance magnitudes and profile shapes, consistent with previous work on elbow assistance \cite{ref6,ref43}. Torque increased using a cubic smoothstep function over the prescribed rise width and was maintained until completion of the lift. The action grid contained 840 profiles.

Participants performed loaded curls in contexts defined by load and fatigue-induction dose. Each participant selected a load comfortable for repeated curls as their high load, with intermediate and low loads 1 and 2 kg lower. Low and high loads were 1 and 3 kg for one participant, 2 and 4 kg for three, and 3 and 5 kg for two. Fatigue-induction dose was defined as $d=n_{\mathrm{pre}}/30$, where $n_{\mathrm{pre}}\in\{0,15,30\}$ was the prescribed number of unassisted curls.
These curls were performed using each participant’s high load before testing at the load assigned to the context.
Dose represented prescribed exercise rather than measured fatigue. The five contexts combined low and high loads with $d=0$ and $d=1$, plus an intermediate-load condition at $d=0.5$ (Fig.~\ref{fig:2}(f)).

Between contexts, participants rested until they reported readiness to begin preparation for the next context. To reduce the potential influence of residual fatigue on lower-dose conditions, the two $d=0$ contexts were tested first, followed by the $d=0.5$ context and then the two $d=1$ contexts. Load order was randomized separately for $d=0$ and $d=1$.

For ordinal assessments, participants judged whether the assistance was helpful during lifting. Each context comprised 26 exposures, each beginning with a single loaded curl, with additional repetitions permitted as needed before feedback. The initial five exposures used predefined profiles common to all contexts, sampling different combinations of torque magnitude, onset position, and rise width. A reference profile was presented first, followed by four profiles in an order randomized for each participant and retained across contexts. For exposures 6 to 25, eligible profiles excluded settings already selected in the current context or assigned to other contexts for that participant, distributing observations across different settings. One eligible profile was sampled uniformly at each exposure, subject to minimum-distance constraints relative to the two preceding settings. Repeating the reference profile at the end gave each profile one comparison as the current setting and one as the preceding setting. Each context yielded 26 ordinal assessments and 25 direct pairwise responses. Detailed sampling constraints and the elbow initial-profile values are provided in Supplementary Section C.

\begin{figure*}[!b]
\centerline{\includegraphics[width=\textwidth]{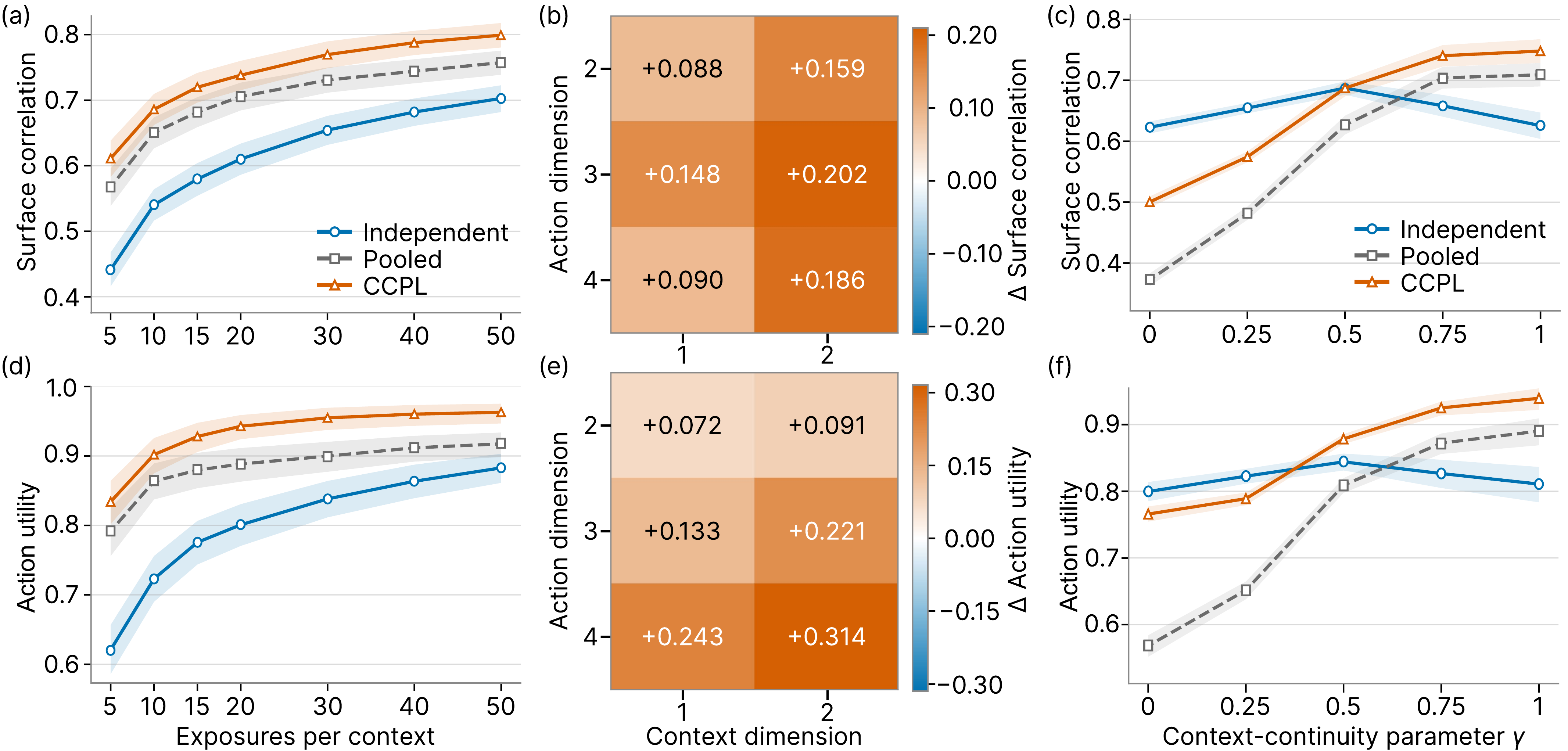}}
\caption{Simulation performance across feedback budgets, dimensionality configurations, and context continuity. Top and bottom rows show reconstruction and preference-based Bayesian optimization (PBO), respectively. Surface correlation is the Pearson correlation between reconstructed and ground-truth utility slices.
Action utility is the normalized ground-truth utility of the recommended action. (a, d) Performance across budgets at $\gamma=1$. (b, e) Mean paired differences (CCPL minus independent learning) at $\gamma=1$ and 10 exposures per context. (c, f) Budget-averaged performance across context-continuity settings $\gamma$, calculated as normalized areas under the curves over 5 to 50 exposures per context. Results were averaged across contexts within each instance and then across 30 instances per configuration. Line plots weight the six dimensionality configurations equally. Shading shows 95\% bootstrap confidence intervals. Complete budget curves appear in Supplementary Fig. S1.
}
\label{fig:3}
\end{figure*}

\subsubsection{Analysis Setup}
\label{sec:human_analysis}
Likelihood and variance settings followed Section~\ref{sec:gp_model}. Action coordinates were normalized to [0,1] using their predefined bounds, and context coordinates using their minima and maxima across each participant's experimental conditions. Within each study, all three strategies shared fixed action length scales, and CCPL additionally used fixed context length scales.
The length-scale values and sensitivity analysis for the context length scales are provided in Supplementary Section C.
All eligible partners were used for ordinal-derived comparisons, without the 32-partner limit applied in simulation.

For each participant and context, a separate GP preference model was fitted to all recorded observations.
Its posterior mean served as the common reference for the three strategies’ limited-feedback reconstructions.

\subsubsection{Evaluation Protocols}

We examined context-related preference structure using Pearson correlations between each participant's full-data reference landscapes for all pairs of contexts. Quantitative comparisons used the complete action grid, including all three elbow action dimensions. For each participant, a least-squares line summarized how reference-landscape correlations changed with Euclidean distance between the normalized context coordinates defined in Section~\ref{sec:human_analysis}. A negative slope indicated lower landscape similarity at greater context distances.

For reconstruction at $K=2,\ldots,20$ exposures per context, all windows of $K$ consecutive exposures were evaluated, with starting positions matched across contexts and strategies for each participant.
Fitting used ordinal assessments within each window and direct responses only when both compared exposures were inside it.
Ordinal-derived comparisons and weights were rebuilt from these observations.
Reconstruction was measured by Pearson correlation between posterior means and the corresponding full-data references over the complete action grid.
Normalization and constant-landscape handling are detailed in Supplementary Section C2.
We used $K=5$ exposures per context as a representative limited-feedback condition.
For each participant, we estimated when independent learning's mean reconstruction correlation first reached CCPL's value at $K=5$ from below, interpolating linearly between adjacent budgets without extrapolation.

Models fitted to the reconstruction windows predicted recorded direct responses for which both compared exposures were outside the training window.
Test responses were matched across strategies within each window but varied with window position and feedback budget.
Predictive probabilities of preferring either action or reporting a tie incorporated posterior uncertainty without min-max normalization, as detailed in Supplementary Section A.
Prediction was evaluated using mean negative log likelihood (NLL), in nats per response.

For source reuse, each context served in turn as the target within each participant.
We evaluated all windows of $K=2,\ldots,20$ consecutive target exposures and all combinations of $N$ other contexts as sources, with $N=1,2$ for ankle and $N=1,\ldots,4$ for elbow.
CCPL and pooled learning used the target window and complete source histories, whereas independent learning used the same target window alone.
Only target exposures counted toward $K$, with source histories treated as already available.
Target reconstruction followed the limited-feedback procedure.

\subsection{Statistical Analysis}

Simulation outcomes were averaged equally across contexts within each instance before calculating paired method differences. Overall estimates weighted the six dimensionality configurations equally. Confidence intervals (CIs) were obtained from 10,000 bootstrap resamples of instances within each configuration, preserving this weighting.

The ankle and elbow studies were analyzed separately, with participants as the statistical units.
Within each participant, reconstruction correlations were averaged equally
across windows and then contexts at each budget.
Held-out NLL was first averaged over eligible responses within each window,
then equally across windows and finally across contexts at each budget.
For source reuse, outcomes were averaged equally over windows, then source configurations and target contexts, at each target budget and source count.
Method differences were paired within participants, with NLL reductions defined as comparator minus CCPL.
Each participant contributed one slope relating reference-landscape similarity to context distance.
Group estimates weighted the six participants equally, with CIs from 10,000 participant-level bootstrap resamples.

Results are reported as means and unstandardized paired differences with percentile 95\% confidence intervals. No null-hypothesis significance tests or p-values were used. Confidence intervals were not adjusted for multiple comparisons.

\section{Results}

\subsection{Simulation Results}

\subsubsection{Limited-Feedback Reconstruction}

At $\gamma=1$, CCPL achieved the highest mean reconstruction correlation at every evaluated feedback budget (Fig.~\ref{fig:3}(a)). With five exposures per context, mean correlations were 0.611 [95\% CI: 0.583, 0.638] for CCPL, 0.441 [0.416, 0.467] for independent learning, and 0.567 [0.538, 0.596] for pooled learning. The mean CCPL advantage over independent learning decreased from 0.170 at five exposures to 0.097 at 50 exposures.

\subsubsection{Preference-Based Bayesian Optimization}

CCPL also achieved the highest mean normalized ground-truth utility of the recommended action across the evaluated budgets at $\gamma=1$ (Fig.~\ref{fig:3}(d)). With five exposures per context, mean recommended-action utilities were 0.834 [95\% CI: 0.802, 0.864] for CCPL, 0.621 [0.586, 0.657] for independent learning, and 0.792 [0.756, 0.826] for pooled learning. As in reconstruction, the mean advantage over independent learning was smaller at 50 than at five exposures.

\subsubsection{Effects of Dimensionality and Context Continuity}

At $\gamma=1$ and 10 exposures per context, CCPL outperformed independent learning in both reconstruction and optimization across all six dimensionality configurations (Fig.~\ref{fig:3}(b), (e), Supplementary Section B2).
For each tested action dimensionality, CCPL's mean advantage in reconstruction and optimization was larger with two context dimensions than with one.
However, sampling more contexts in the two-dimensional setting also provided more observations for sharing, which may have contributed to the larger gains.
As action dimensionality increased, the optimization advantage grew, whereas the reconstruction advantage did not increase monotonically.

CCPL showed negative transfer relative to independent learning at $\gamma=0$ and $0.25$ for both reconstruction and optimization, assessed by normalized area under the feedback-budget curve (Fig.~\ref{fig:3}(c), (f)). At $\gamma=0$, the paired difference (CCPL minus independent learning) was \textminus{}0.123 [95\% CI: \textminus{}0.129, \textminus{}0.116] for reconstruction and \textminus{}0.034 [\textminus{}0.046, \textminus{}0.021] for optimization. At $\gamma=0.5$, the reconstruction difference was 0.000 [\textminus{}0.006, 0.006], whereas optimization favored CCPL by 0.034 [0.023, 0.046]. At $\gamma=1$, the differences were positive for both reconstruction, 0.122 [0.115, 0.130], and optimization, 0.129 [0.108, 0.151].

In contrast, CCPL had higher budget-averaged performance than pooled learning at all five $\gamma$ settings, with paired-difference confidence intervals above zero for both reconstruction and optimization. These differences ranged from 0.036 to 0.128 for reconstruction and from 0.049 to 0.198 for optimization.

\subsection{Human Results}

\subsubsection{Context-Related Preference Structure}

\begin{figure}[!t]
\centering
\includegraphics[width=\columnwidth]{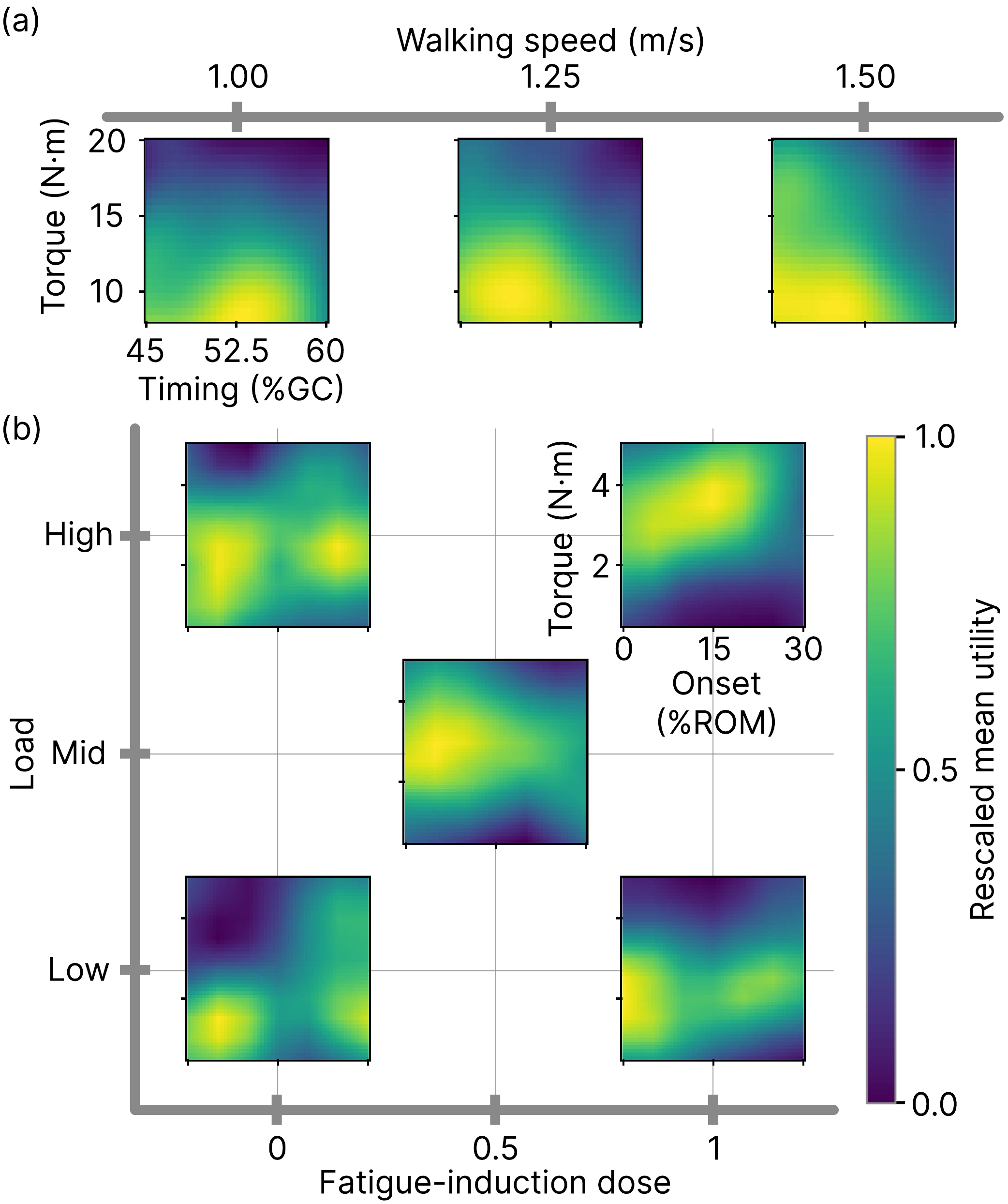}
\caption{Full-data reference utility landscapes across operating contexts. (a) Ankle landscapes over peak timing and torque at three walking speeds. (b) Elbow projections over onset position and torque at five load and fatigue-induction-dose conditions. Dose d denotes prescribed unassisted curls divided by 30. References were fitted and normalized separately for each participant and context, with elbow projections taking the maximum over rise width. Colors show means across six participants per study, rescaled to [0,1] within each context. GC and ROM denote gait cycle and calibrated elbow range of motion. Visualization details and individual landscapes appear in Supplementary Section C1 and Figs. S2 and S3.}
\label{fig:4}
\end{figure}

Participant-averaged full-data reference landscapes showed differences in the location and extent of high-utility regions across contexts (Fig.~\ref{fig:4}). For ankle assistance, high utility spanned a broader range of peak timings at lower torques and a narrower range at higher torques. With increasing walking speed, these regions shifted toward earlier peak timings and higher torque magnitudes. For elbow assistance, the projections showed two high-utility regions at early and late onsets at $d=0$. Higher loads favored higher torques; within each load, the dominant peak shifted toward higher torques and earlier onsets from $d=0$ to $d=1$. A single dominant region appeared at intermediate load with $d=0.5$ and high load with $d=1$.

Within participants, full-data reference-landscape similarity decreased with normalized context distance on average. The mean participant-level slope relating reference-landscape similarity to normalized context distance was \textminus{}0.424 [95\% CI: \textminus{}0.726, \textminus{}0.209] in the ankle study and \textminus{}0.325 [\textminus{}0.529, \textminus{}0.129] in the elbow study (Supplementary Fig. S4). Slopes were negative for all six participants in each study, although the relationship was nearly flat for one elbow participant.

\subsubsection{Limited-Feedback Reconstruction}

\begin{figure*}[!t]
\centerline{\includegraphics[width=\textwidth]{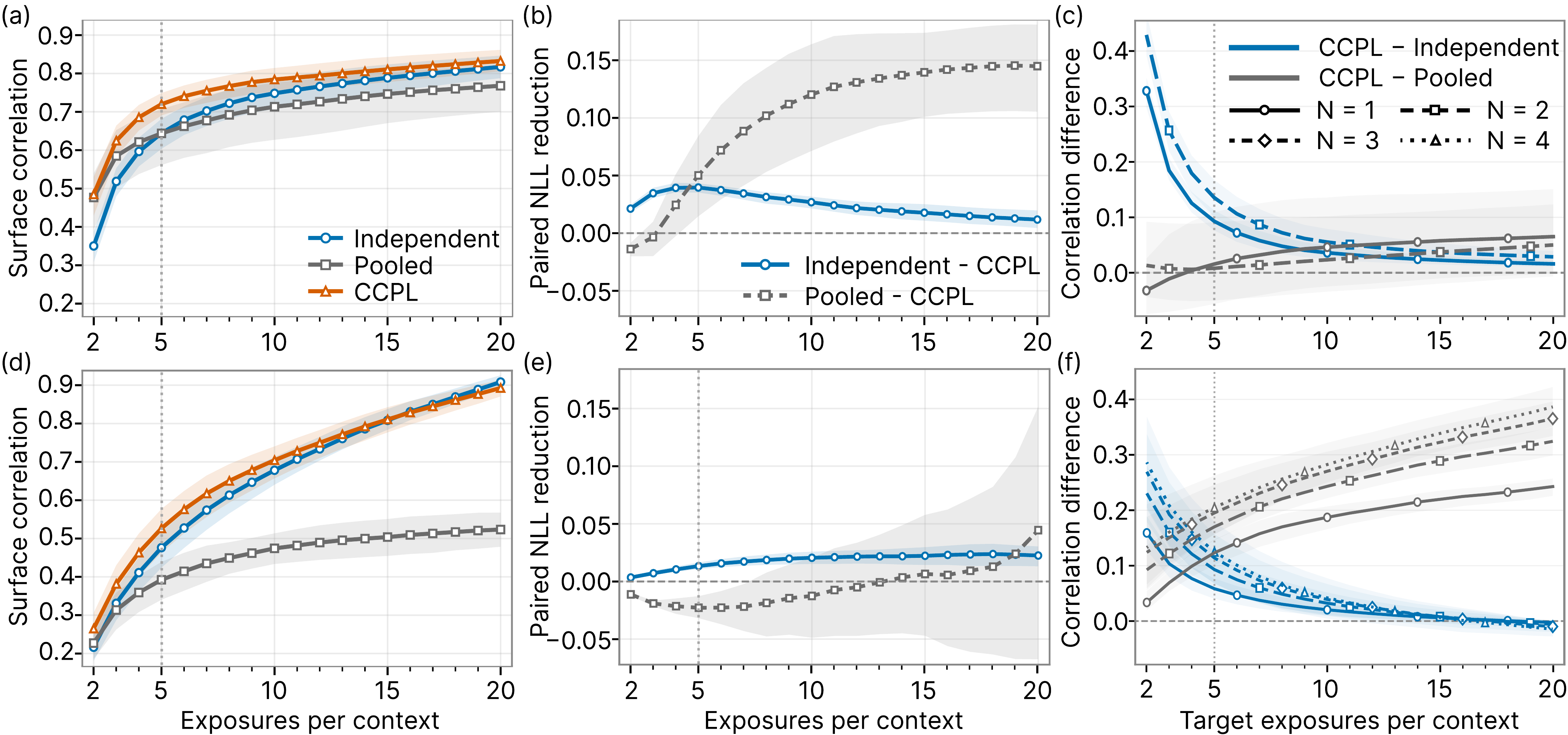}}
\caption{Human-study evaluation of preference learning. Top and bottom rows show ankle and elbow results, respectively. (a, d) Reconstruction Pearson correlations with full-data references at $K=2,\ldots,20$ exposures per context. (b, e) Paired reductions in held-out negative log likelihood (NLL), calculated as independent or pooled learning minus CCPL in nats per response. Models from (a, d) predicted responses comparing two exposures outside each training window. Test sets were matched across methods but varied across windows and budgets. (c, f) Paired target-reconstruction correlation differences, calculated as CCPL minus independent or pooled learning, using $K$ target exposures and $N$ source contexts. CCPL and pooled learning used full source histories excluded from $K$, while independent learning used target observations alone. Line styles and markers distinguish $N=1,2$ for ankle and $N=1,\ldots,4$ for elbow. Curves show means across six participants per study. Shading shows pointwise 95\% percentile bootstrap confidence intervals from participant-level resampling. Vertical dotted lines mark $K=5$, and horizontal dashed lines mark zero. Positive differences favor CCPL.}
\label{fig:5}
\end{figure*}

With five exposures per context, CCPL achieved the highest mean correlations with the full-data references, at 0.720 for ankle and 0.526 for elbow (Fig.~\ref{fig:5}(a), (d)). Ankle participant-paired CCPL differences were 0.076 [95\% CI: 0.064, 0.089] relative to independent learning and 0.076 [0.031, 0.137] relative to pooled learning. The corresponding elbow differences were 0.050 [0.030, 0.070] and 0.134 [0.118, 0.154], respectively. After averaging across contexts and windows, all six participants in each study showed higher correlation with CCPL than with independent learning.

The reconstruction advantage over independent learning was smaller at larger feedback budgets (Fig.~\ref{fig:5}(a), (d), Supplementary Fig. S5). At 20 exposures per context, the mean ankle difference was 0.0153 [95\% CI: 0.0088, 0.0217]. In the elbow study, the mean difference became negative at 16 exposures and reached \textminus{}0.0156 [\textminus{}0.0294, \textminus{}0.0008] at 20 exposures. At 20 exposures per context, the evaluated windows covered 19.8\% of the ankle reference history (20/101) and 76.9\% of the elbow history (20/26).

Independent learning was estimated to require 7.94 [95\% CI: 7.34, 8.63] exposures per context for ankle and 6.02 [5.57, 6.47] for elbow to match the reconstruction performance achieved by CCPL with five exposures. These estimates correspond to approximately 37\% fewer exposures for ankle and 17\% fewer for elbow with CCPL.

\subsubsection{Held-Out Response Prediction}

CCPL achieved lower mean held-out NLL than independent learning at every evaluated feedback budget in both studies (Fig.~\ref{fig:5}(b), (e)). The 95\% confidence intervals for the participant-paired reductions were above zero. At five exposures per context, mean CCPL NLL was 0.810 for ankle and 1.006 for elbow. The NLL reductions relative to independent learning were 0.040 [95\% CI: 0.035, 0.043] for ankle and 0.013 [0.011, 0.017] for elbow, with lower NLL for CCPL in all six participants in each study. At 20 exposures, the corresponding reductions were 0.012 [0.004, 0.020] and 0.023 [0.013, 0.030]. In the elbow study, this prediction advantage persisted despite CCPL's lower reconstruction correlation relative to independent learning.

Comparisons with pooled learning differed between studies. At five exposures, CCPL reduced ankle NLL by 0.050 [95\% CI: 0.015, 0.084], whereas elbow NLL was higher with CCPL by 0.023 [0.013, 0.032]. At 20 exposures, the mean elbow NLL reduction relative to pooled learning was 0.044 [\textminus{}0.067, 0.152], with a confidence interval that included zero.

\subsubsection{Reuse of Source Observations}

With five target exposures and full histories from all other contexts, mean reconstruction correlations for CCPL were 0.780 for ankle and 0.601 for elbow. Ankle participant-paired differences were 0.135 [95\% CI: 0.101, 0.167] relative to independent learning and 0.008 [\textminus{}0.050, 0.091] relative to pooled learning (Fig.~\ref{fig:5}(c)). The corresponding elbow differences were 0.125 [0.068, 0.179] and 0.205 [0.156, 0.262], respectively (Fig.~\ref{fig:5}(f)). At this budget with all sources, all six participants in each study showed higher reconstruction correlation with CCPL than with independent learning.

The advantage over independent learning decreased as target feedback increased (Fig.~\ref{fig:5}(c), (f)). With all sources at 20 target exposures, the mean difference was 0.0287 [95\% CI: 0.0179, 0.0390] for ankle and \textminus{}0.0144 [\textminus{}0.0277, 0.0004] for elbow. Adding a second source increased mean CCPL correlation throughout the evaluated budgets in the ankle study. In the elbow study, additional sources improved mean correlation at smaller budgets, but this pattern reversed at larger budgets. At 20 target exposures, mean elbow correlation decreased from 0.906 with one source to 0.894 with four sources. Confidence intervals for the CCPL differences from pooled learning included zero in every evaluated ankle condition and were above zero in every elbow condition (Fig.~\ref{fig:5}(c), (f)).

\section{Discussion}

This study examined whether a user's preference landscape varies smoothly across operating conditions and when this continuity supports learning from limited feedback. In simulations, sharing observations through CCPL improved reconstruction and PBO relative to independent learning when preferences varied smoothly, but led to negative transfer when continuity was weak (Fig.~\ref{fig:3}(c), (f)). The ankle and elbow studies provided evidence consistent with context continuity, as full-data reference landscapes fitted separately for each participant and context tended to be more similar between nearby conditions. This relationship suggests that observations from one condition may be informative about the same user's preferences in another. In both studies, CCPL achieved higher reconstruction correlations with these references under limited feedback and lower held-out NLL across the evaluated feedback budgets than independent learning. Together, these results support context continuity as a basis for sharing preference observations under limited feedback, with benefits depending on how preferences vary across conditions.

CCPL’s reconstruction gains over independent learning were larger with fewer target observations in both the limited-feedback and source-reuse analyses. With more target feedback, the mean advantage in agreement with the full-data references diminished and became negative in the elbow study. This attenuation may partly reflect the fact that the same exposure count represented a larger fraction of the reference history in the elbow study than in the ankle study. Despite the diminished reconstruction advantage, CCPL retained its held-out prediction advantage over independent learning. This emphasis on limited feedback is practically important because extensive feedback collection in each operating condition is constrained by time, physical effort, and fatigue \cite{ref11}. The gains with limited target feedback therefore suggest that reusing observations from related conditions could reduce the new feedback needed to personalize assistance for the same user.

Unlike pooled learning, CCPL shares observations while retaining context-specific preference estimates, consistent with contextual and multitask learning approaches \cite{ref33,ref36,ref37}. In simulations, CCPL achieved higher budget-averaged performance than pooled learning at all tested levels of context continuity, including levels at which it underperformed independent learning. This indicates that retaining context can improve upon treating observations from different conditions as interchangeable, although it does not by itself prevent negative transfer when context continuity is weak. In both human studies, CCPL improved reconstruction over pooled learning under limited feedback, but this advantage was less consistent in source reuse and held-out response prediction. Comparison with independent learning assesses whether sharing helps, whereas comparison with pooled learning assesses whether retaining context adds value when observations are shared.

The simulation results suggest that context sharing can support assistance selection even when its reconstruction advantage is limited. At $\gamma=0.5$, the budget-averaged reconstruction difference from independent learning was close to zero, whereas budget-averaged recommended-action utility was higher with CCPL (Fig.~\ref{fig:3}(c), (f)). This contrast is plausible because identifying a high-utility action does not require accurate reconstruction of every part of the landscape. PBO also selected subsequent actions using each model, whereas reconstruction used action sequences selected without preference feedback. Sharing could therefore affect both the regions explored and the action ultimately recommended. These findings motivate testing CCPL in online personalization as users encounter changing operating conditions \cite{ref15,ref44}. Related work has used experience replay to maintain gait-phase estimation across locomotor conditions \cite{ref45}. For preference-based personalization, an optimizer using CCPL could retain observations from previous contexts and use posterior utility and uncertainty to choose the next assistance setting for the user to experience and evaluate.

A further consideration for online personalization is that preference feedback may depend on the assistance experienced over time. For example, in the ankle study, experiencing higher-torque assistance at unfavorable peak timings may have reduced participants’ subsequent preference for higher-torque settings, potentially contributing to the broader high-utility timing range at lower torques in the reference landscapes (Fig.~\ref{fig:4}(a)). Such experience dependence is consistent with reports that users adapt to exoskeleton assistance over time \cite{ref3} and that preferred assistance can change with device exposure \cite{ref23}. More broadly, online personalization may need to consider human–robot co-adaptation, in which the assistance policy and user behavior adapt through interaction \cite{ref46}. Extending CCPL to account for recent interactions \cite{ref47} and changes over time \cite{ref48,ref49}, with appropriate discounting of older observations \cite{ref50}, could help determine how earlier feedback should contribute to estimates of current preferences.

CCPL could also be extended across users and to personalization objectives beyond preference.
Transfer learning has supported personalization by adapting models trained on other individuals using limited data from a new user \cite{ref51, ref52}.
In exoskeleton personalization, experienced wearers' exploration histories have accelerated preference-based optimization for inexperienced wearers \cite{ref53}.
Combining such transfer with CCPL could support sharing across users and contexts, provided that individual differences are preserved.
The other direction is context-aware sharing for physiological or task-performance objectives \cite{ref5, ref54}.
This would require adapting the observation model to these measurements and assessing whether the responses vary smoothly across operating contexts.

Several limitations should be considered when interpreting these findings. Generalizability is limited by the small cohort of healthy adults and the range of operating contexts evaluated. Larger cohorts are needed to examine how context continuity and the benefits of sharing vary across individuals. In the elbow study, fatigue-induction dose represented prescribed exercise rather than measured fatigue, and the fixed ordering of lower-dose conditions prevented separating dose effects from effects of testing order. Additional limitations concern the evaluation design. Because preferences may change with experience, we compared the learning strategies retrospectively using the same recorded exposure and response histories. Although this design provided a common basis for comparison, it did not evaluate online assistance selection in humans. Because the full-data references were model estimates, higher reconstruction correlations do not necessarily indicate more accurate recovery of users' underlying preferences. The held-out analysis provided a complementary evaluation against recorded responses excluded from training. Finally, CCPL was evaluated using context variables predefined for the walking and lifting tasks. Identifying suitable variables may become more difficult as the range of tasks expands. Future work could therefore examine whether context representations derived from human movement capture preference similarity across tasks and provide a useful basis for sharing.

\section{Conclusion}

This study supports context continuity as a basis for sharing a user's preference observations across operating conditions.
Simulations showed reconstruction and PBO gains over independent learning when preferences varied smoothly, but negative transfer when continuity was weak.
In retrospective analyses of ankle and elbow exoskeleton data, nearby conditions tended to have more similar reference landscapes.
CCPL improved reconstruction and held-out prediction relative to independent learning under limited feedback, while benefits over pooled learning varied.
Reconstruction gains, including those from source reuse, were larger with fewer target observations.
Prospective human studies are needed to determine whether these learning gains improve assistance selection or reduce required exposures during online personalization.

\section*{Acknowledgment}
The authors would like to thank all participants for their time and effort towards this study.

\clearpage
\includepdf[pages=-,pagecommand={\thispagestyle{empty}}]{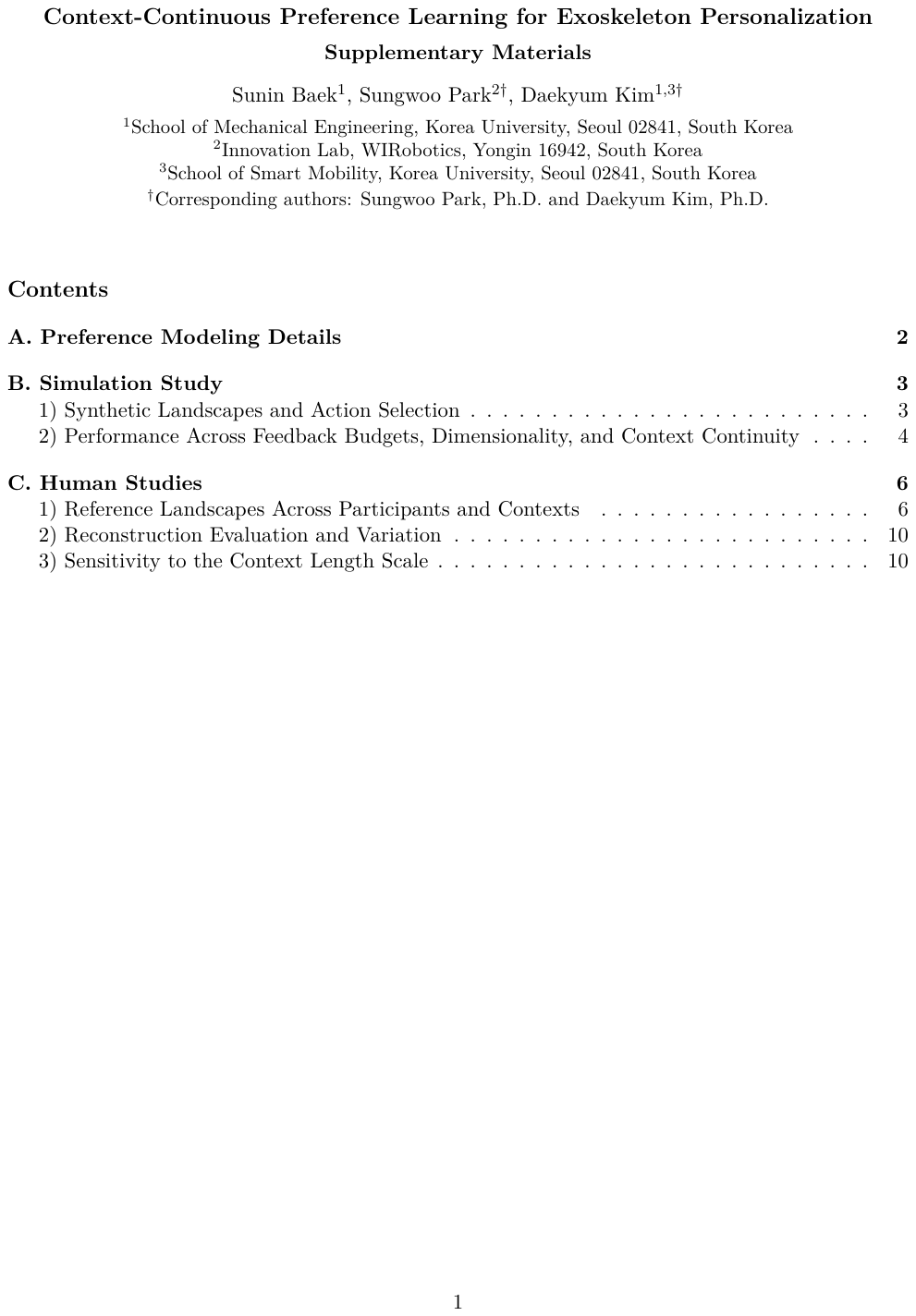}

\end{document}